\documentclass[letterpaper]{article} % DO NOT CHANGE THIS
\usepackage{aaai2026}  % DO NOT CHANGE THIS% 
\usepackage{times}  % DO NOT CHANGE THIS
\usepackage{helvet}  % DO NOT CHANGE THIS
\usepackage{courier}  % DO NOT CHANGE THIS
\usepackage[hyphens]{url}  % DO NOT CHANGE THIS
\usepackage{graphicx} % DO NOT CHANGE THIS
\usepackage{natbib}  % DO NOT CHANGE THIS AND DO NOT ADD ANY OPTIONS TO IT
\usepackage{caption} % DO NOT CHANGE THIS AND DO NOT ADD ANY OPTIONS TO IT
\usepackage{algorithm}
\usepackage{algorithmic}
\usepackage{amsfonts}
\usepackage{amsmath}
\usepackage{amssymb}
\usepackage{array}
\usepackage{booktabs}
\usepackage{caption}
\usepackage{colortbl}
\usepackage{enumitem}
\usepackage{graphicx}
\usepackage{listings}
\usepackage{longtable}
\usepackage{mdframed}
\usepackage{multirow}
\usepackage{multicol}
\usepackage{newfloat}
\usepackage{subcaption}
\usepackage{tikz}
\usepackage{xspace}

\definecolor{lightgray2}{gray}{0.95}
\newcommand{\graybgline}{\cellcolor{lightgray2}}

\definecolor{myred}{RGB}{237,28,80 }
\definecolor{scarlet}{RGB}{255,36,0}
\definecolor{keywordred}{RGB}{200,34,34}
\definecolor{iceblue}{RGB}{214, 230, 245}
\definecolor{creamyellow}{RGB}{255,246,213}
\definecolor{mygreen}{RGB}{112,166,100}

\newcommand{\ourmethod}{SentiDetect\xspace}
\newcommand{\ourmethodTT}{\textit{\textbf{\fontfamily{lmtt}\selectfont \ourmethod}}\xspace}
\newcommand{\ourmethodC}{\textit{\textbf{\fontfamily{lmtt}\selectfont \ourmethod-SDC}}\xspace}
\newcommand{\ourmethodP}{\textit{\textbf{\fontfamily{lmtt}\selectfont \ourmethod-SDP}}\xspace}

\newcommand{\GPTThreeFive}{GPT-3.5-Turbo\xspace}
\newcommand{\GPTThreeFiveTT}{\textbf{\fontfamily{lmtt}\selectfont GPT-3.5-Turbo}\xspace}
\newcommand{\Gemini}{Gemini-1.5-Pro\xspace}
\newcommand{\GeminiTT}{\textbf{\fontfamily{lmtt}\selectfont Gemini-1.5-Pro}\xspace}
\newcommand{\GPTFour}{GPT-4-0613\xspace}
\newcommand{\GPTFourTT}{\textbf{\fontfamily{lmtt}\selectfont GPT-4-0613}\xspace}
\newcommand{\LLaMa}{LLaMa-3.3\xspace}
\newcommand{\LLaMaTT}{\textbf{\fontfamily{lmtt}\selectfont LLaMa-3.3}\xspace}

\newcommand{\ClaudeTT}{\textbf{\fontfamily{lmtt}\selectfont Claude-3}\xspace}

\usepackage{newfloat}
\usepackage{listings}
\DeclareCaptionStyle{ruled}{labelfont=normalfont,labelsep=colon,strut=off} % DO NOT CHANGE THIS
\floatstyle{ruled}
\newfloat{listing}{tb}{lst}{}
\floatname{listing}{Listing}
\title{Model-Agnostic Sentiment Distribution Stability Analysis for Robust LLM-Generated Texts Detection}
\author{
    Siyuan Li\textsuperscript{\rm 1}, Xi Lin\textsuperscript{\rm 1}, Guangyan Li\textsuperscript{\rm 2}, Zehao Liu\textsuperscript{\rm 1}, Aodu Wulianghai\textsuperscript{\rm 1}, Li Ding\textsuperscript{\rm 1}, Jun Wu\textsuperscript{\rm 1}, Jianhua Li\textsuperscript{\rm 1}
}
\affiliations{
    \textsuperscript{\rm 1}School of Computer Science, Shanghai Jiao Tong University, Shanghai, China \\
    \textsuperscript{\rm 2}Institute of Automation, Chinese Academy of Sciences, Beijing, China\\
    \{siyuanli, linxi234, melusine.wlhad, liding, junwuhn, lijh888\}@sjtu.edu.cn, liuzehao.cn@gmail.com, liguangyan2022@ia.ac.cn
}

\nocopyright

\usepackage{bibentry}
\begin{document}

\maketitle

\begin{abstract}
The rapid advancement of large language models (LLMs) has resulted in increasingly sophisticated AI-generated content, posing significant challenges in distinguishing LLM-generated text from human-written language.
Existing detection methods, primarily based on lexical heuristics or fine-tuned classifiers, often suffer from limited generalizability and are vulnerable to paraphrasing, adversarial perturbations, and cross-domain shifts.
In this work, we propose \ourmethodTT, a model-agnostic framework for detecting LLM-generated text by analyzing the divergence in sentiment distribution stability.
Our method is motivated by the empirical observation that LLM outputs tend to exhibit emotionally consistent patterns, whereas human-written texts display greater emotional variability.
To capture this phenomenon, we define two complementary metrics: sentiment distribution consistency and sentiment distribution preservation, which quantify stability under sentiment-altering and semantic-preserving transformations.
We evaluate \ourmethod on five diverse datasets and a range of advanced LLMs, including \GeminiTT, \ClaudeTT, \GPTFourTT, and \LLaMaTT.
Experimental results demonstrate its superiority over state-of-the-art baselines, with over 16\% and 11\% F1 score improvements on \Gemini and \GPTFour, respectively.
Moreover, \ourmethod also shows greater robustness to paraphrasing, adversarial attacks, and text length variations, outperforming existing detectors in challenging scenarios.
\end{abstract}

% Uncomment the following to link to your code, datasets, an extended version or similar.
% You must keep this block between (not within) the abstract and the main body of the paper.
% \begin{links}
%     \link{Code}{https://aaai.org/example/code}
%     \link{Datasets}{https://aaai.org/example/datasets}
%     \link{Extended version}{https://aaai.org/example/extended-version}
% \end{links}

\section{Introduction}
\begin{figure}[!t]
    \centering
    \includegraphics[width=\linewidth]{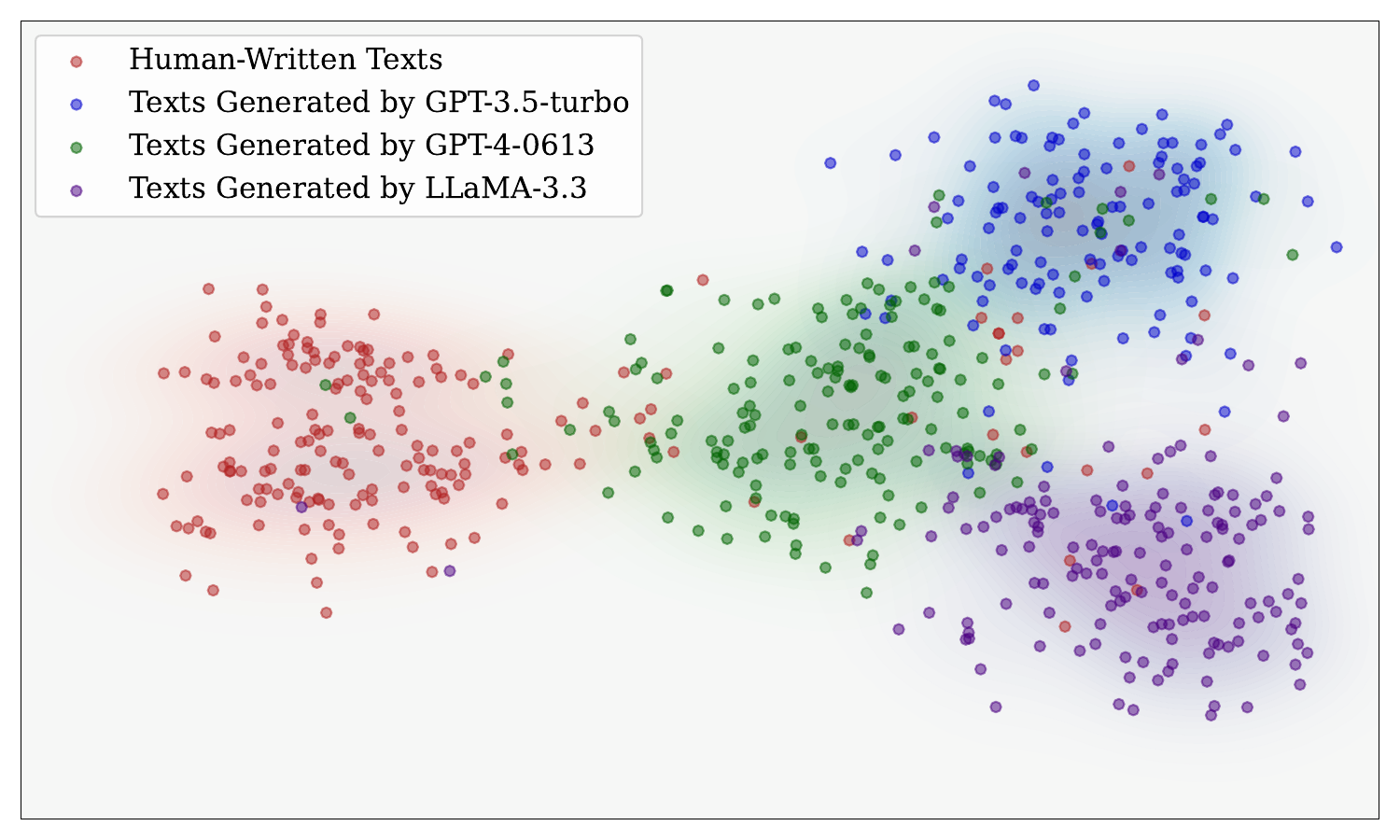}
    \caption{UMAP projection illustration of \textit{\textcolor{keywordred}{\textbf{sentiment distribution stability features for LLM-generated texts and human-written}}} in the Review dataset. 
    Each point represents an embedded paragraph ($\leq$ 64 tokens) encoded with Sentiment Distribution Consistency features.
    \textit{\textbf{LLM-generated texts exhibit consistent sentiment patterns, forming distinguishable clusters from human-written texts}}. }
    \label{figure:scatter-plot}
\end{figure}

Advanced large language models (LLMs) have achieved remarkable proficiency in processing and generating natural language across diverse domains, from logical reasoning to creative writing~\cite{yang2023harnessing}.
As these powerful generative models become increasingly integrated into writing workflows~\cite{sun2024trustllm, lei2025pald}, their outputs now permeate various forms of digital content, including news articles, academic papers, and social platforms~\cite{yuan2022wordcraft}.
However, LLM-generated content still poses numerous security risks and potential misuses, such as phishing and the creation of fake news, which can lead to fraud and factual inaccuracies~\cite{lee2023language, sadasivan2023can}.
Furthermore, the growing sophistication of LLM-generated text presents new challenges in distinguishing machine-generated texts from human-written texts~\cite{chen2025online}.

The emergence of various challenges has sparked considerable research into developing \textit{automatic LLM-generated text detection approaches}.
To address the misuse of LLMs, there are three types of LLM-generated text detectors: watermark~\cite{kirchenbauer2023reliability}, classification~\cite{solaiman2019release}, and statistic-based detectors~\cite{tulchinskii2024intrinsic}.
Although watermark-based detectors can identify embedded watermarks in the output text, they are susceptible to paraphrasing~\cite{krishna2024paraphrasing} and can be bypassed by open-source LLMs~\cite{touvron2023llama}.
Supervised classification is effective within specific domains but struggles with text from new sources or unfamiliar LLMs~\cite{pu2023deepfake}.
Existing statistic-based detectors analyze statistical features of texts to determine their origins and obtain comparable performance to supervised detectors.
However, these methods require access to the source LLM, which is often unfeasible, and current detectors remain vulnerable to paraphrasing attacks~\cite{krishna2024paraphrasing}.

In contrast to prior approaches, we investigate a different perspective: \textit{\textbf{(RQ) can sentiment patterns embedded in language be leveraged for authorship attribution without supervised training or access to model parameters?}}
Motivated by this question, we conduct empirical studies and make a key observation: the stability of emotional expressions in text emerges as a strong and distinctive indicator for attributing authorship.
Building on this insight, we propose the hypothesis that LLM-generated content tends to exhibit relatively stable sentiment distributions under specific ``low-emotional" transformations, irrespective of textual genre or stylistic variations~\cite{hu2023survey, vijay2025neutral}. 
This phenomenon likely stems from the nature of LLM training, which involves exposure to vast, uniformly curated corpora~\cite{naveed2023comprehensive, shen2024donowcharacterizingevaluating}.
Figure~\ref{figure:scatter-plot} demonstrates how this feature can effectively distinguish LLM-generated text from human-written texts.
These subtle yet measurable differences in sentiment fluctuation offer a previously underexplored fingerprint for distinguishing LLM-generated from human-written text.

Building on this insight, we present \ourmethodTT, a novel model-agnostic framework that detects LLM-generated text by analyzing divergence in sentiment distribution stability.
Motivated by the observation that LLM-generated content tends to exhibit more emotionally consistent patterns, \ourmethod quantifies this distinction through two complementary unsupervised metrics: sentiment distribution consistency and sentiment distribution preservation.
These metrics capture stability under sentiment-altering and semantic-preserving transformations, enabling relatively interpretable detection without access to model internals or fine-tuned data.
Comprehensive experiments across five datasets and multiple advanced LLMs demonstrate that \ourmethod significantly outperforms existing baselines in both accuracy and robustness, particularly under paraphrasing, adversarial, and length-variant settings.

The contributions are summarized as follows: 
\begin{itemize}
    \item \textit{\textbf{First-of-its-kind study focusing on sentiment analysis for LLM-generated text.}} \ourmethodTT is a model-agnostic framework that identifies LLM-generated text through the divergence in sentiment distribution stability, requiring neither model access nor supervised training.
    \item \textbf{\textit{Strong zero-shot performance on various LLMs and textual domains.}} Extensive evaluations are conducted on \GeminiTT, \LLaMaTT, \GPTThreeFiveTT, and \GPTFourTT.
    Besides, \ourmethod improves F1 score on five datasets, including news, code, essays, papers, and community comments by up to 22.61\%.
    \item \textbf{\textit{In-depth analysis of detection robustness.}} We provide an in-depth analysis of the robustness against various adversarial tactics, including paraphrasing, perturbation, and text length across various datasets and LLMs.
\end{itemize}

\section{Related Works}
\subsection{LLM-Generated Text Detection}
Current LLM-generated text detection approaches can be categorized into three paradigms: supervised classification, watermarking techniques,  and statistical analysis methods~\cite{chakraborty2024position}.

\textbf{Supervised Training Detectors.}
Classification-based methods treat the detection as a binary prediction task. 
Early works like Grover~\cite{zellers2019defending} generate articles from titles, and OpenAI employed a fine-tuned RoBERTa model to classify GPT-2 outputs, noting larger models produce better text~\cite{solaiman2019release}.
Furthermore, adversarial learning can also be introduced to train the robust detector~\cite{hu2024radar}.
However, these supervised methods require extensive labeled datasets and constant updates to keep pace with evolving generative models~\cite{soto2024few, zhang2024detecting}.

\textbf{Watermarking Detectors.}
Watermark-based approaches embed identifiable patterns during text generation. 
Initial work included adversarial watermarking transformers for binary message encoding~\cite{watermark0:abdelnabi2021adversarial}, with recent advances proposing probability distribution perturbations for more robust marking~\cite{Watermark1-forLLM:kirchenbauer2023watermark}. 
Despite effective watermarks resisting removal without degrading text quality, these methods struggle with conditional text generation~\cite{watermark2:fu2023watermarking} and are vulnerable to paraphrasing~\cite{krishna2024paraphrasing}.

\textbf{Statistical Analysis Detectors.}
Statistical-based detectors~\cite{bao2024fast} identify abnormal values in text features like entropy and n-gram frequency. 
OpenAI's detector uses the logarithmic probability of the model~\cite{solaiman2019release}, while GLTR employs entropy and probability rank. 
Recently, DetectGPT demonstrated that LLM-generated text often falls in the negative curvature region of the LLM's logarithmic probability function, resulting in a curvature-based classifier~\cite{mitchell2023detectgpt}.
Subsequent studies have developed a faster version of this method~\cite{bao2024fast}.
These methods provide valuable insights into the distinctive features of LLM-generated text without supervised training.

\begin{figure*}[!t]
    \centering
    \includegraphics[width=\linewidth]{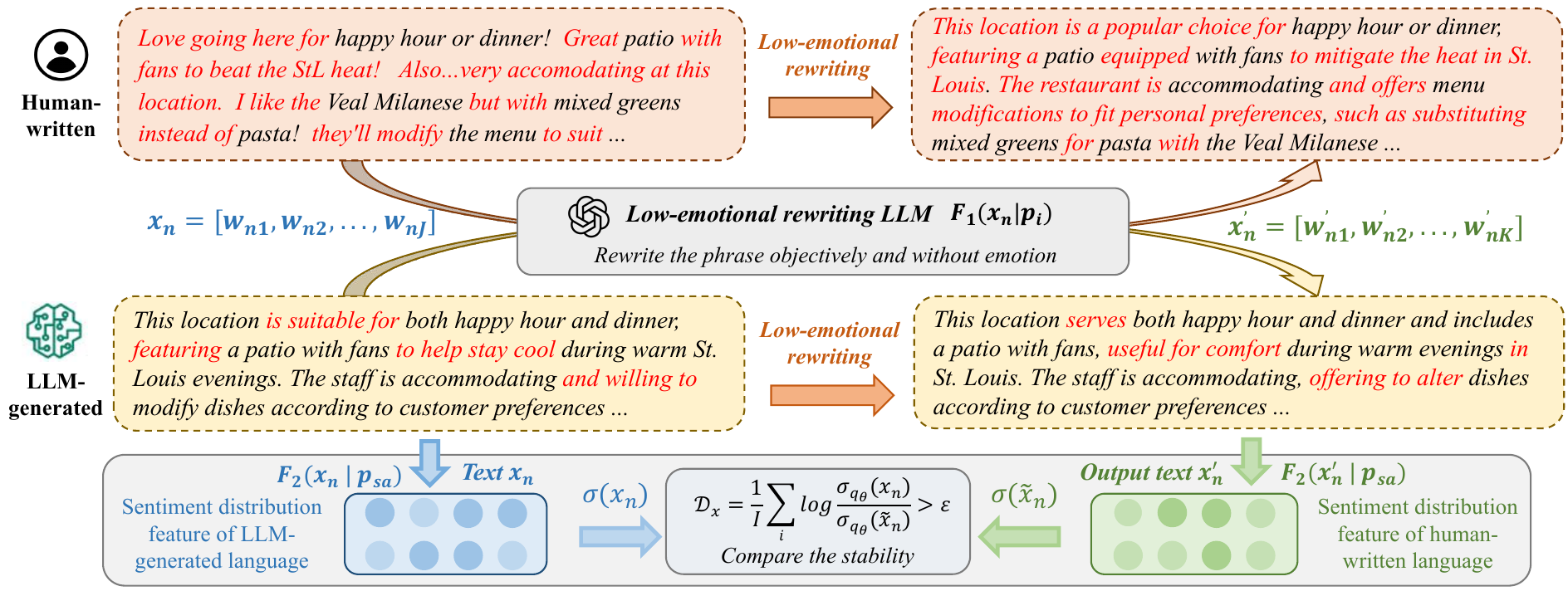}
    \caption{Illustration of the proposed \ourmethodTT. Detecting LLM-generated text by sentiment distribution stability analysis through low-emotional rewriting.}
    \label{figure:framework}
\end{figure*}
\subsection{Sentiment Analysis in the Era of LLMs}
Recent research has expanded sentiment analysis beyond basic classification to more nuanced examination of textual sentiment patterns, such as aspect-based or multifaceted subjective analysis~\cite{zhang2022survey}.
The understanding ability of LLM has a significant influence on sentiment analysis~\cite{zhong2023can}.
Notably, researchers have explored innovative methods to leverage generative models for sentiment analysis tasks. 
Deng et al. developed a semi-supervised framework that utilizes language models to generate sentiment labels for training smaller models~\cite{deng2023llms}, highlighting the potential of LLMs in analyzing sentiment features.
Besides, LLMs consistently outperform smaller models for zero-shot sentiment analysis of diverse text corpora~\cite{zhang2024detecting}.
Therefore, for the sentiment stability analysis considered in this work, we designed a pipeline combined with LLMs to implement the idea.

\section{Methodology}
\subsection{LLM-Generated Text Detection Definition}
Given the probability measures $\mathcal{P}$ and $\mathcal{Q_\theta}$ on metric space $\mathcal{X}$, suppose we have a set of IID candidate texts $\{x_n\}_{n=1}^{N}$, which are either from the human-written distribution $\mathcal{P}$ or LLM-generated distribution $\mathcal{Q}_\theta$. 
Our goal is to determine whether each $x_n$ originates from the $\mathcal{Q}_{\theta}$ of source $q_\theta$: 
\begin{equation} 
    \hat{y}_n = \underset{ y_n \in \{y_\mathcal{P}, y_{\mathcal{Q}_\theta} \} }{\operatorname{argmax}} P (y_n \mid x_n, N, \mathcal{P}, \mathcal{Q_\theta}).
\end{equation}
Here, $x_n= [w_{n, 1}, w_{n, 2}, \ldots, w_{n, J} ]$ denotes the $n$-th text sequence, $w_{n,j}$ represent the $j$-th word in $x_{n}$. 
When $N=1$, the paradigm collapses into a single-instance detection task.

\textbf{Overview of \ourmethod.}
An overview of the designed detection pipeline is shown in Figure~\ref{figure:framework}, since LLMs are less effective at directly conducting complex sentiment feature analysis. 
The pipeline comprises three stages: 
\textit{(I) Low-emotional rewriting (LER)}: Using zero-shot LLMs to rewrite the text under "low-emotional" instructions; 
\textit{(II) Sentiment feature extraction}: Extracting sentiment distribution features from both original and rewritten texts; 
\textit{(III) Stability divergence analysis}: Measuring the difference in stability of sentiment distributions to identify the input text.

\subsection{Sentiment Distribution Stability Analysis}
While LLM-generated text often appears contextually appropriate, its sentiment distribution patterns exhibit subtle yet measurable differences from human-written text. 
Our finding reveals this key phenomenon as follows:
\begin{mdframed}[backgroundcolor=mygreen!10, linewidth=1pt, linecolor=mygreen!50, skipabove=0.8\baselineskip, skipbelow=0.8\baselineskip]
    \textbf{\textcolor{keywordred}{Divergence in low-emotional stability}}: 
    \textit{LLM-generated text \textbf{maintains consistent sentiment distribution patterns} during low-emotional rewriting, while human-written text exhibits significant shifts in sentiment expression patterns.}
\end{mdframed}
Besides, to operationalize this hypothesis, we implement the detection pipeline as illustrated in Figure~\ref{figure:framework}, since LLMs are empirically less effective at directly conducting complex sentiment feature analysis.
First, we introduce a low-emotional but semantic-preserving rewriting function $F_1(\cdot \mid p_i)$ to strategically modifies sentiment pattern of the text $x_n$. 
To be specific, $F_1(\cdot \mid p_i)$ maps text $x_{n}$ to $x^{\prime}_{n}= [w_{n,1}^{\prime}, w_{n,2}^{\prime}, \ldots, w_{n,K}^{\prime}]$ of indefinite length.
We repeat the above process for each $p_i$ to ensure robust pattern capture.

Subsequently, the sentiment distribution analyzer $F_2(\cdot \mid p_{sa})$ is employed to quantify the sentiment distribution feature of both the original text $x_n$ and its rewritten versions $x^{\prime}$, as shown in Figure~\ref{figure:framework}. 
The function $F_2(\cdot)$ maps each input text $x$ to a fixed-length vector representation $F_2(x \mid p_{sa})$, referred to as the sentiment distribution pattern. 
To construct this vector, we apply $F_2$ across semantic-preserving LERs of the original text, denoted as $x^{\prime}_1, \dots, x^{\prime}_I$. 
This process yields $I$ sentiment distributions, which are then concatenated to form a unified feature embedding. 
This embedding serves as a stable representation for measuring divergence across different samples.
Specifically, the $F_2(\cdot \mid p_{sa})$ in the experiments is implemented using a standard 3-class sentiment classifier, which outputs a probability vector over the classes (negative, neutral, positive). 
This level of granularity empirically captures patterns of sentiment stability while avoiding the ambiguity introduced by overly fine-grained labels.
The sentiment distributions of all the rewritten versions are concatenated, resulting in a fixed-length signal that enables a reliable comparison between the samples.

The sentiment distribution pattern of text $x$ is denoted as $\sigma(x) = F_2(x \mid p_{sa})$. 
By comparing the sentiment distribution before and after LER, we can quantify the distribution divergence as $\log(\sigma_{q_\theta}(x)) - \log(\sigma_{q_\theta}(x^{\prime}))$, highlighting differences in sentiment stability between LLM outputs and human-written text. 
The parameters $p_i$ and $p_{sa}$ represent carefully designed transformation prompts.
With the help of the concepts of functions $F_1(x \mid p_i)$ and $F_2(x \mid p_{sa})$, we define the distribution stability divergence in sentiment features as follows:
\begin{equation}
\begin{aligned}
     \mathcal{D}(x, q_\theta) & \triangleq \log \sigma(x) - \mathbb{E}_{x^{\prime} \sim F_1(x \mid p_i)} \log \sigma(x^{\prime}).
\end{aligned} 
\end{equation}
This is referred to as the sentiment distribution stability divergence hypothesis, which describes the characteristic differences between LLM-generated and human-written texts.

\subsection{LLM-Generated Text Detection via Sentiment Distribution Stability Divergence}
Based on the empirical hypothesis, LLM-generated texts and human-written text can be distinguished by the change of sentiment distribution feature:
\begin{equation}
    \mathcal{D}_x =\frac{1}{I} \sum\limits_i \log (\frac{\sigma_{q_\theta}(x)}{\sigma_{q_\theta}(F_1(x \mid p_i))}).
\end{equation}
The input $x$ will be identified as LLM-generated text when the divergence $\mathcal{D}_x < \varepsilon$, where $\varepsilon$ is the decision threshold.

\textbf{LER Rewriting Instructions.}
For the implementation of the sentiment distribution stability analysis, we design a set of transformation instructions that guide language models to modify the emotional tone of a text while preserving its semantic content. 
For example, instructions such as ``\textit{Decrease the emotional intensity while preserving semantic content}" or ``\textit{Rewrite this paragraph objectively}" serve to elicit controlled variations in sentiment.
We formally define each instruction as $p_i$, drawn from templates like ``\textit{Rewrite objectively}" or ``\textit{Use a machine-like tone}", and optimize them using AutoPrompt~\cite{ma2024fairness} to ensure effectiveness and consistency.
These transformation instructions are simple in form yet central to our method: they induce minimal changes in LLM-generated text due to the inherent regularity of its sentiment structure, while often triggering more noticeable sentiment shifts in human-written text.
This contrast serves as a key signal for distinguishing between LLM-generated and human-written text.

To capture this phenomenon, we implement \ourmethod by leveraging two key properties of sentiment distribution stability: \textit{sentiment distribution consistency (SDC)} and \textit{sentiment distribution preservation (SDP)}.

\textbf{Sentiment Distribution Consistency.} Language models show minimal changes in sentiment distributions when transforming their own generated text, as autoregressive models tend to produce content with stable pattern features.
Therefore, we use the low-emotional invariance between output and input to measure the sentiment distribution recognition of LLM towards a given input text $x$.
We use prompt $p_i$ to prompt LLM $q_\theta$ to perform LER transformation on text $x$. 
If the text $x$ is generated from LLM, \textit{the degree of change from output to input should be small}.
We quantify the sentiment distribution consistency through:
\begin{equation}
\begin{aligned}
    \text{SDC}(x) & = \mathbb{E}_{p_i} \left[ \left\| \log \sigma(x), \log \sigma(x') \right\|_1 \right] \\
    & = \frac{1}{I} \sum_{i=1}^{I} ( \log \sigma(x) - \log \sigma(F_1(x \mid p_i)) ).
\end{aligned}
\end{equation}
For the instruction $p_i$, we present some of them:
\begin{itemize}[itemsep=1pt, topsep=1.5pt]
    \item \textit{Please rewrite this more straightforwardly.}
    \item \textit{Polish this in a machine-like objective tone.}
\end{itemize}
We employ automated methods to optimize the instruction $p_i$~\cite{ma2024fairness}.

\textbf{Sentiment Distribution Preservation.}
In addition, we give the definition of the other key property, i.e. sentiment distribution preservation.
Suppose we perform sentiment-independent and inverse transformations (such as converting people or extensions and abbreviations) on the texts. 
In that case, it will \textit{output with the same sentiment distribution as the input text in the previous}. 
We refer to the property that LLM-generated text maintains stable sentiment distributions under semantic-preserving transformations as sentiment distribution preservation.
Besides, we elaborate instructions to represent a pair of inverse mapping pairs $\mathcal{F}(\cdot)$ and $\mathcal{F}^{-1}(\cdot)$.
$\mathcal{F}(\cdot)$ represents requesting LLM by forward transformation hint words to produce a transformed text output, while $\mathcal{F}^{-1}(\cdot)$ represents its opposite hint word.
We can use the following indicators to measure sentiment distribution preservation~\cite{mao2024detecting}:
\begin{equation}
    \text{SDP}(x) = \| \log \sigma(x), \log \sigma( \mathcal{F}^{-1}(\mathcal{F}(F_1(x \mid p_i))) ) \|_1.
\end{equation}
By mapping twice through instructions with opposite meanings, the LLM-generated text preserves its original sentiment distribution feature. 

\textbf{Detection on short texts.} At the same time, we observe that this zero-shot detection method requires sufficient text length for reliable analysis.
For example, suppose we have two short text samples:
\begin{itemize}[itemsep=1pt, topsep=1.5pt]
    \item LLM-generated text: \textit{I feel very happy today because it's sunny. I plan to take a walk in the park and enjoy the beauty of nature.}
    \item Human-written text: \textit{Feeling extra good today with the sun shining brightly. Planning to stroll in the park and breathe in some fresh air.}
\end{itemize}
Notably, LLM-generated short texts can closely mimic human-written content, making them harder to distinguish based on limited context. 
For shorter samples, LLMs can closely approximate human-like sentiment expressions, making detection more challenging. 
However, this limitation can be mitigated by aggregating multiple short texts to reveal stable distributional patterns, a direction further explored in our experiments on input text length.

\begin{table*}[!th]
    \centering
    \small
    \setlength{\tabcolsep}{7pt}
    \resizebox{\linewidth}{!}{
    \begin{tabular}{l||ccccc|c}
    \hline \hline
    \rowcolor{gray!40}
    \textbf{Methods} & \textbf{News} & \textbf{HumanEval} & \textbf{Student Essay} & \textbf{Paper} & \textbf{Yelp Review} & \textbf{Avg} \\
    \hline
    \multicolumn{7}{c}{\graybgline \textbf{\GPTFourTT}} \\
    \hline
    LogRank (GPT-2) & 41.50$\pm$1.72 & 56.08$\pm$2.14 & 42.86$\pm$1.35 & 55.48$\pm$1.89 & 58.46$\pm$2.03 & 50.88$\pm$1.83 \\
    \rowcolor{gray!10} RoBERTa-base~\cite{liu2019roberta} & 42.14$\pm$2.25 & 39.51$\pm$0.87 & 36.40$\pm$1.62 & 39.45$\pm$1.03 & 37.40$\pm$0.95 & 38.98$\pm$1.34 \\
    RoBERTa-large~\cite{liu2019roberta} & 43.47$\pm$1.84 & 54.94$\pm$2.21 & 45.03$\pm$1.57 & 49.26$\pm$1.42 & 63.27$\pm$2.35 & 51.19$\pm$1.88 \\
    \rowcolor{gray!10} GPT-Zero~\cite{GPTZero} & 52.68$\pm$0.97 & 53.43$\pm$1.45 & 38.60$\pm$2.12 & 71.21$\pm$2.24 & 69.20$\pm$2.17 & 57.02$\pm$1.79 \\
    DetectGPT~\cite{mitchell2023detectgpt} & 29.47$\pm$0.85 & 48.53$\pm$1.63 & 48.09$\pm$1.54 & 61.20$\pm$1.96 & 53.16$\pm$1.72 & 48.09$\pm$1.54 \\
    \rowcolor{gray!10} Ghostbuster~\cite{verma2024ghostbuster} & 46.85$\pm$1.33 & 35.84$\pm$0.92 & 52.77$\pm$0.93 & 62.38$\pm$2.12 & 62.68$\pm$2.08 & 52.10$\pm$1.48 \\
    RAIDAR~\cite{mao2024detecting} & 54.52$\pm$1.88 & 55.20$\pm$1.76 & 53.16$\pm$1.65 & 75.28$\pm$2.41 & 68.74$\pm$2.22 & 61.38$\pm$1.98 \\
    \rowcolor{gray!10} Binoculars~\cite{hans2024spotting} & \underline{58.02}$\pm$0.91 & 67.48$\pm$1.17 & 51.60$\pm$1.04 & 78.23$\pm$1.88 & 62.76$\pm$1.76 & 63.62$\pm$1.35 \\
    R-Detect~\cite{song2025deep} & 56.18$\pm$1.63 & 63.41$\pm$1.73 & 54.52$\pm$0.94 & 76.17$\pm$1.72 & 67.18$\pm$1.43 & 63.49$\pm$1.49 \\
    
    \hline
    \rowcolor{creamyellow!50} 
    \ourmethodC (Ours) & 
    \textbf{59.03}$_{\textcolor{black}{\uparrow1.01}}$$\pm$1.92 &
    \underline{78.28}$_{\textcolor{black}{\uparrow10.8}}$$\pm$2.33 &
    \textbf{58.83}$_{\textcolor{black}{\uparrow4.31}}$$\pm$1.81 &
    \textbf{82.80}$_{\textcolor{black}{\uparrow4.57}}$$\pm$1.45 &
    \underline{77.02}$_{\textcolor{black}{\uparrow7.82}}$$\pm$2.28 &
    \textbf{71.19}$_{\textcolor{black}{\uparrow7.57}}$$\pm$1.96 \\
    
    \rowcolor{creamyellow!50}
    \ourmethodP (Ours) & 56.73$\pm$1.75 &
    \textbf{82.74}$_{\textcolor{black}{\uparrow15.26}}$$\pm$2.39 & 
    \underline{55.61}$_{\textcolor{black}{\uparrow1.09}}$$\pm$1.68 &
    \underline{79.20}$_{\textcolor{black}{\uparrow0.97}}$$\pm$2.31 &
    \textbf{80.16}$_{\textcolor{black}{\uparrow10.96}}$$\pm$1.42 &
    \underline{70.89}$_{\textcolor{black}{\uparrow7.27}}$$\pm$1.91 \\
    
    \hline
    \multicolumn{7}{c}{\graybgline \textbf{ \GeminiTT } } \\
    \hline
    LogRank (GPT-2) & 44.79$\pm$3.15 & 53.71$\pm$1.91 & 45.19$\pm$3.24 & 46.25$\pm$3.59 & 55.78$\pm$2.34 & 48.72$\pm$3.06 \\
    \rowcolor{gray!10} RoBERTa-base~\cite{liu2019roberta} & 43.52$\pm$2.89 & 49.93$\pm$0.68 & 33.76$\pm$2.65 & 39.45$\pm$1.40 & 44.71$\pm$2.50 & 41.94$\pm$2.02 \\
    RoBERTa-large~\cite{liu2019roberta} & 43.15$\pm$1.22 & 55.35$\pm$2.60 & 32.83$\pm$2.85 & 47.24$\pm$2.66 & 58.35$\pm$3.48 & 51.33$\pm$2.73 \\
    \rowcolor{gray!10} GPT-Zero~\cite{GPTZero} & 46.24$\pm$2.49 & 46.92$\pm$3.51 & 39.46$\pm$1.19 & 66.92$\pm$3.12 & 62.50$\pm$1.83 & 55.98$\pm$2.57 \\
    DetectGPT~\cite{mitchell2023detectgpt} & 41.67$\pm$2.98 & 64.88$\pm$1.60 & 47.55$\pm$2.32 & 66.64$\pm$3.52 & 63.76$\pm$0.60 & 59.41$\pm$2.29 \\
    \rowcolor{gray!10} Ghostbuster~\cite{verma2024ghostbuster} & 44.31$\pm$0.80 & 40.53$\pm$1.51 & 39.97$\pm$2.22 & 52.72$\pm$0.32 & 48.69$\pm$6.13 & 46.77$\pm$2.36 \\
    RAIDAR~\cite{mao2024detecting} & 45.84$\pm$4.43 & 71.64$\pm$5.09 & 51.94$\pm$5.64 & 68.72$\pm$4.72 & 69.14$\pm$4.83 & 61.53$\pm$5.15 \\
    \rowcolor{gray!10} Binoculars~\cite{hans2024spotting} & \underline{52.24}$\pm$1.63 & 60.31$\pm$1.82 & 50.63$\pm$1.28 & 70.29$\pm$1.10 & 58.95$\pm$1.50 & 58.48$\pm$1.47\\
    R-Detect~\cite{song2025deep} & 49.08$\pm$1.11 & 57.38$\pm$1.69 & 53.12$\pm$1.97 & 68.95$\pm$1.39 & 67.43$\pm$0.97 & 59.19$\pm$1.43 \\
    
    % \midrule
    \hline
    \rowcolor{creamyellow!50} 
    \ourmethodC (Ours) & 
    \textbf{54.93}$_{\textcolor{black}{\uparrow2.69}}$$\pm$1.04 & 
    \underline{77.34}$_{\textcolor{black}{\uparrow5.70}}$$\pm$2.91 & 
    \textbf{61.85}$_{\textcolor{black}{\uparrow8.73}}$$\pm$2.50 &
    \underline{71.55}$_{\textcolor{black}{\uparrow1.26}}$$\pm$0.83 & 
    \textbf{79.19}$_{\textcolor{black}{\uparrow10.05}}$$\pm$3.20 &
    \textbf{71.65}$_{\textcolor{black}{\uparrow10.12}}$$\pm$2.10 \\
    
    \rowcolor{creamyellow!50} 
    \ourmethodP (Ours) & 
    50.93$_{\textcolor{black}{\uparrow4.69}}$$\pm$2.16 &
    \textbf{80.88}$_{\textcolor{black}{\uparrow9.24}}$$\pm$2.71 &
    \underline{54.29}$_{\textcolor{black}{\uparrow1.17}}$$\pm$1.69 &
    \textbf{72.35}$_{\textcolor{black}{\uparrow2.06}}$$\pm$2.07 & 
    \underline{78.03}$_{\textcolor{black}{\uparrow8.89}}$$\pm$0.68 &
    \underline{70.39}$_{\textcolor{black}{\uparrow8.86}}$$\pm$2.29 \\
    \hline
    \end{tabular}
    }
    \caption{Detection F1 scores of the content generated from \GPTFour and \Gemini for News, HumanEval, Student Essay, Paper, and Yelp Review datasets. 
    The \textbf{best} and the \underline{second-best} scores are in bold and underlined, respectively.}
    \label{table:GPT4andGemini}
\end{table*}

\section{Experiments}
\subsection{Experimental Setup}
\subsubsection{Datasets.}
We evaluate our method on five datasets, following~\cite{verma2024ghostbuster, mao2024detecting}: \textbf{News} contains 5,000 real news articles from 50 journalists and corresponding AI-generated versions, produced via a two-stage title-to-article process. 
\textbf{HumanEval} includes 164 programming tasks with signatures, docstrings, and test cases, covering reasoning, algorithms, and math. 
\textbf{Student Essay} comprises high school and university-level papers from the BAWE corpus along with LLM-generated counterparts. 
\textbf{Yelp Review} features 1,000 human-written Yelp reviews and machine-generated versions of similar length. 
\textbf{Paper Abstract} consists of 500 ACL 2023–2024 abstracts, with synthetic versions created from the first 15 words using LLMs, ensuring zero-shot validity by avoiding overlap with pretraining data.

\subsubsection{Baselines.}
We compare our method with several state-of-the-art LLM-generated text detectors: \textbf{GPTZero}~\cite{GPTZero} is a leading commercial detector for texts from models like \textit{ChatGPT}, \textit{GPT-4}, and \textit{LLaMa};
\textbf{DetectGPT}~\cite{mitchell2023detectgpt} identifies GPT-generated text using probability curvature in a zero-shot setting without supervised training;
\textbf{Ghostbuster}~\cite{verma2024ghostbuster} detects texts through directly using a series of language models; 
\textbf{RAIDAR}~\cite{mao2024detecting} improves accuracy through rewriting-based analysis that captures text modification patterns;
\textbf{Binoculars}~\cite{hans2024spotting} is a zero-shot detection method through contrasting predictions from a pair of pre-trained LLMs;
\textbf{R-Detect}~\cite{song2025deep} employs a non-parametric kernel relative test to compare the distributional closeness of a sample to text corpora.

Notably, although combining SDC and SDP into a unified metric has been explored, the performance gain was marginal and reduced interpretability. 
Thus, we report them separately for clarity. 
We also verify the semantic fidelity of inverse mapping pairs using BLEURT~\cite{sellam2020bleurt}, observing high cross-domain consistency to support reliable SDP analysis.

\subsection{Main Detection Performance}
\begin{figure}[!t]
    \centering
    \includegraphics[width=\linewidth]{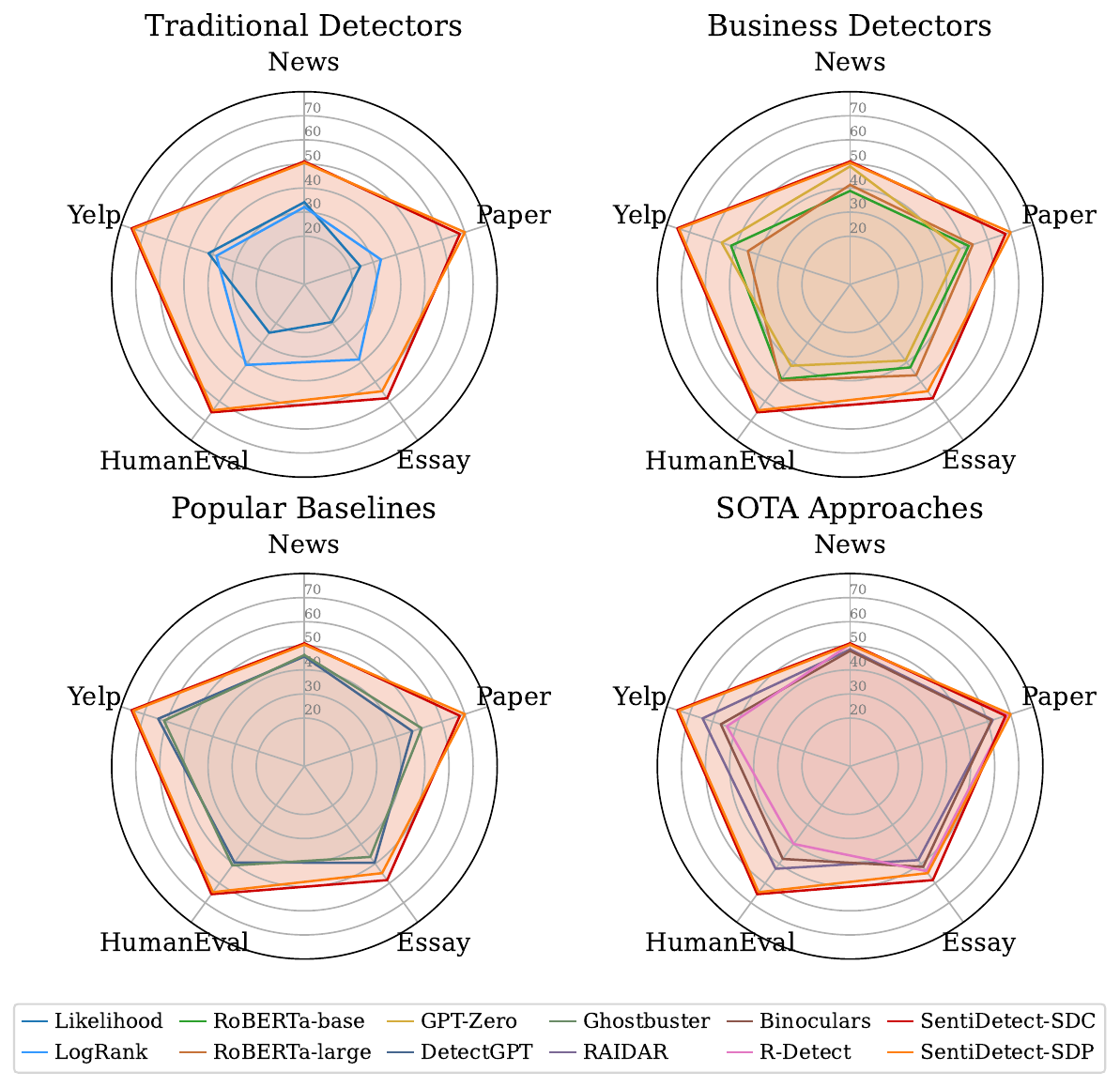}
    \caption{Visualized comparison between \ourmethod with four categories of baselines (traditional detectors, business detectors, popular baselines, and SOTA approaches) on five datasets, evaluated on content from \ClaudeTT.
    }
    \label{figure:datasets}
\end{figure}

\begin{table*}[!t]
    \centering
    \small    
    % \footnotesize
    \setlength{\tabcolsep}{7pt}
    \resizebox{\linewidth}{!}{
    \begin{tabular}{l||ccccc|c} 
    \hline
    \hline
    \rowcolor{gray!40}
    \textbf{Method}
    & \textbf{News} & \textbf{HumanEval} & \textbf{Student Essay} & \textbf{Paper} & \textbf{Yelp Review} & \textbf{Avg} \\
    \hline
    \multicolumn{7}{c}{\graybgline \textbf{\LLaMaTT} } \\
    \hline
    LogRank (GPT-2) & 58.27$\pm$2.50 & 69.38$\pm$3.05 & 54.37$\pm$1.85 & 68.39$\pm$0.66 & 72.25$\pm$1.81 & 67.07$\pm$1.97 \\
    \rowcolor{gray!10} RoBERTa-base~\cite{liu2019roberta} & 58.45$\pm$0.79 & 61.30$\pm$2.97 & 40.72$\pm$0.91 & 73.60$\pm$3.19 & 49.91$\pm$2.60 & 60.82$\pm$2.09 \\
    RoBERTa-large~\cite{liu2019roberta} & 54.06$\pm$0.93 & 72.71$\pm$3.71 & 38.48$\pm$1.55 & 63.62$\pm$2.12 & 64.02$\pm$1.30 & 63.60$\pm$1.92 \\
    \rowcolor{gray!10} GPT-Zero~\cite{GPTZero} & 55.83$\pm$2.82 & 68.49$\pm$2.91 & 51.78$\pm$1.64 & 87.61$\pm$2.38 & 77.56$\pm$3.67 & 72.37$\pm$2.68 \\
    DetectGPT~\cite{mitchell2023detectgpt} & 57.42$\pm$3.12 & 76.70$\pm$0.78 & 59.13$\pm$2.81 & 70.38$\pm$3.81 & 72.48$\pm$2.77 & 69.25$\pm$2.06 \\
    \rowcolor{gray!10} Ghostbuster~\cite{verma2024ghostbuster} & 57.36$\pm$3.28 & 50.02$\pm$0.85 & 50.72$\pm$1.43 & 67.78$\pm$3.23 & 65.42$\pm$3.66 & 60.21$\pm$2.49 \\
    RAIDAR~\cite{mao2024detecting} & 61.35$\pm$3.29 & 81.06$\pm$3.41 & 62.56$\pm$2.63 & 79.04$\pm$1.82 & 75.06$\pm$4.27 & 71.81$\pm$3.08 \\
    \rowcolor{gray!10} Binoculars~\cite{hans2024spotting} & 65.80$\pm$0.98 & 67.82$\pm$1.31 & 58.32$\pm$1.72 & 83.64$\pm$1.93 & 58.28$\pm$1.29 & 66.77$\pm$1.45 \\
    R-Detect~\cite{song2025deep} & 67.73$\pm$1.25 & 59.38$\pm$1.10 & 62.40$\pm$1.51 & 81.77$\pm$1.07 & 63.27$\pm$1.29 & 66.91$\pm$1.24 \\
    
    \hline
    \rowcolor{creamyellow!50} 
    \ourmethodC (Ours) & 
    \textbf{70.17}$_{\textcolor{black}{\uparrow2.44}}$$\pm$0.87 & 
    \textbf{95.69}$_{\textcolor{black}{\uparrow14.63}}$$\pm$2.05 &
    \textbf{72.39}$_{\textcolor{black}{\uparrow9.83}}$$\pm$2.02 &
    \textbf{89.22}$_{\textcolor{black}{\uparrow5.58}}$$\pm$0.94 & \underline{85.61}$_{\textcolor{black}{\uparrow10.55}}$$\pm$2.68 & 
    \textbf{85.17}$_{\textcolor{black}{\uparrow13.36}}$$\pm$1.71 \\
    
    \rowcolor{creamyellow!50} 
    \ourmethodP (Ours) & 
    \underline{68.28}$_{\textcolor{black}{\uparrow0.55}}$$\pm$0.82 & 
    \underline{90.01}$_{\textcolor{black}{\uparrow8.95}}$$\pm$1.84 & 
    \underline{68.96}$_{\textcolor{black}{\uparrow6.40}}$$\pm$1.70 & 
    \underline{87.98}$_{\textcolor{black}{\uparrow4.34}}$$\pm$3.08 & 
    \textbf{89.05}$_{\textcolor{black}{\uparrow13.99}}$$\pm$3.45 & 
    \underline{83.84}$_{\textcolor{black}{\uparrow12.03}}$$\pm$2.18 \\
    \hline
    \end{tabular}
    }
    \caption{Detection F1 scores of the content generated from \LLaMa for News, HumanEval, Student Essay, Paper, and Yelp Review datasets. 
    The \textbf{best} and the \underline{second-best} scores are in bold and underlined, respectively.}
    \label{table:llama}
\end{table*}

\subsubsection{(I) Commercial LLMs: Robust Detection Across Popular LLMs.}
We conduct detection experiments at the paragraph level and compare the results with existing baselines. 
The experimental results on \GPTFour and \Gemini are shown in Table~\ref{table:GPT4andGemini}. 
For processing sentiment distribution information in the input text, we employ sentiment distribution feature analysis using \GPTFour as $F_1(\cdot)$ and $F_2(\cdot)$. 
Our experiments cover multiple datasets, including News, HumanEval, Student Essays, Paper Abstracts, and Review datasets. 
\ourmethod shows consistently superior performance on all datasets, proving its universality in detecting different text types. 
It demonstrates significant advantages in all experimental scenarios, outperforming all baselines on \GPTFour by at least 11.90\% and 11.43\%, respectively. 
When tested on \Gemini, the performance shows no significant decrease compared to \GPTFour, confirming our method's effectiveness even with more powerful LLMs.

\subsubsection{(II) Open-Source LLMs: Detecting Texts from \LLaMa.}
Table~\ref{table:llama} highlights the superior performance of \ourmethod in detecting AI-generated content from open-source \LLaMa. 
These results demonstrate its ability to capture underlying generation patterns regardless of domain or complexity. 
This performance is particularly significant given the architectural and training disparities between open-source models and commercial counterparts. 
Compared with existing methods, \ourmethod delivers an average 13-point F1 improvement, confirming its adaptability to the traits of open-source LLMs. 

\begin{table*}[!th]
    \centering
    \small
    \resizebox{\linewidth}{!}{
    \begin{tabular}{l||ccccc}
    \hline
    \hline
    \rowcolor{gray!40}
    \textbf{Method} & \textbf{News} & \textbf{HumanEval} & \textbf{Student Essay} & \textbf{Paper} & \textbf{Yelp Review} \\
    \hline \hline
    LogRank & 11.61 $\mid$ 52.24 $\mid$ $\downarrow$77.8\% & 7.04 $\mid$ 63.25 $\mid$ $\downarrow$88.9\% & 12.42 $\mid$ 53.30 $\mid$ $\downarrow$76.7\% & 14.10 $\mid$ 61.24 $\mid$ $\downarrow$77.0\% & 22.49 $\mid$ 66.54 $\mid$ $\downarrow$66.2\% \\

    \rowcolor{gray!10} RoBERTa-base~\cite{liu2019roberta} & 10.49 $\mid$ 50.47 $\mid$ $\downarrow$79.2\% & 11.10 $\mid$ 47.72 $\mid$ $\downarrow$76.8\% & 9.43 $\mid$ 43.15 $\mid$ $\downarrow$78.2\% & 16.73 $\mid$ 53.59 $\mid$ $\downarrow$68.8\% & 7.85 $\mid$ 48.87 $\mid$ $\downarrow$83.9\% \\
    
    RoBERTa-large~\cite{liu2019roberta} & 10.42 $\mid$ 51.48 $\mid$ $\downarrow$79.8\% & 18.02 $\mid$ 67.92 $\mid$ $\downarrow$73.5\% & 12.06 $\mid$ 45.58 $\mid$ $\downarrow$73.6\% & 16.65 $\mid$ 63.20 $\mid$ $\downarrow$73.7\% & 16.77 $\mid$ 72.70 $\mid$ $\downarrow$76.9\% \\
    
    \rowcolor{gray!10} GPT-Zero~\cite{GPTZero} & 17.42 $\mid$ 54.99 $\mid$ $\downarrow$68.3\% & 14.55 $\mid$ 58.54 $\mid$ $\downarrow$75.2\% & 19.22 $\mid$ 48.66 $\mid$ $\downarrow$60.5\% & 16.56 $\mid$ 75.21 $\mid$ $\downarrow$78.0\% & 15.57 $\mid$ 78.23 $\mid$ $\downarrow$80.2\% \\
    
    DetectGPT~\cite{mitchell2023detectgpt} & 14.20 $\mid$ 41.62 $\mid$ $\downarrow$65.9\% & 18.60 $\mid$ 70.50 $\mid$ $\downarrow$73.7\% & 22.63 $\mid$ 57.33 $\mid$ $\downarrow$60.5\% & 28.47 $\mid$ 67.98 $\mid$ $\downarrow$58.1\% & 19.43 $\mid$ 68.60 $\mid$ $\downarrow$71.7\% \\

    \rowcolor{gray!10} Ghostbuster~\cite{verma2024ghostbuster} & 15.26 $\mid$ 53.80 $\mid$ $\downarrow$71.7\% & 12.46 $\mid$ 43.08 $\mid$ $\downarrow$71.1\% & 15.48 $\mid$ 49.83 $\mid$ $\downarrow$68.9\% & 14.69 $\mid$ 65.63 $\mid$ $\downarrow$77.6\% & 13.50 $\mid$ 66.86 $\mid$ $\downarrow$79.8\% \\
    
    RAIDAR~\cite{mao2024detecting} & 18.03 $\mid$ 50.28 $\mid$ $\downarrow$64.1\% & 24.26 $\mid$ 71.86 $\mid$ $\downarrow$66.3\% & 24.62 $\mid$ 51.56 $\mid$ $\downarrow$52.2\% & 22.26 $\mid$ 70.40 $\mid$ $\downarrow$68.4\% & 24.61 $\mid$ 65.21 $\mid$ $\downarrow$62.3\% \\

    \rowcolor{gray!10} Binoculars~\cite{hans2024spotting} & 20.37 $\mid$ 57.42 $\mid$ $\boldsymbol{\downarrow}$64.5\% & 20.70 $\mid$ 67.91 $\mid$ $\boldsymbol{\downarrow}$69.5\% & 21.54 $\mid$ 57.38 $\mid$ $\boldsymbol{\downarrow}$62.5\% & 31.33 $\mid$ 82.39 $\mid$ $\boldsymbol{\downarrow}$62.0\% & 25.17 $\mid$ 60.31 $\mid$ $\boldsymbol{\downarrow}$58.3\% \\ 
    
    R-Detect~\cite{song2025deep} & 18.38 $\mid$ 59.92 $\mid$ $\boldsymbol{\downarrow}$69.3\% & 22.51 $\mid$ 70.62 $\mid$ $\boldsymbol{\downarrow}$68.13\% & 19.25 $\mid$ 55.71 $\mid$ $\boldsymbol{\downarrow}$65.4\% & 27.42 $\mid$ 73.60 $\mid$ $\boldsymbol{\downarrow}$62.7\% & 24.27 $\mid$ 66.33 $\mid$ $\boldsymbol{\downarrow}$63.4\% \\ 
    
    \hline
    \rowcolor{creamyellow!50}
    \ourmethodC (Ours)
    & \textbf{32.22} $\mid$ 63.48 $\mid$ $\downarrow$\textbf{49.2\%} 
    & \textbf{48.05} $\mid$ 89.53 $\mid$ $\downarrow$\textbf{46.3\%} 
    & \textbf{37.03} $\mid$ 70.72 $\mid$ $\downarrow$\textbf{47.6\%} 
    & \textbf{46.05} $\mid$ 86.54 $\mid$ $\downarrow$\textbf{46.8\%}
    & \textbf{45.31} $\mid$ 86.40 $\mid$ $\downarrow$\underline{59.2\%} \\

    \rowcolor{creamyellow!50}
    \ourmethodP (Ours) 
    & \underline{28.60} $\mid$ 60.16 $\mid$ $\downarrow$\underline{52.5\%} 
    & \underline{46.79} $\mid$ 89.79 $\mid$ $\downarrow$\underline{47.9\%} 
    & \underline{32.39} $\mid$ 63.42 $\mid$ $\downarrow$\underline{48.9\%} 
    & \underline{40.28} $\mid$ 83.59 $\mid$ $\downarrow$\underline{51.8\%}
    & \underline{43.53} $\mid$ 88.29 $\mid$ $\downarrow$\textbf{50.7\%} \\
    \hline
    \end{tabular}
    }
    \caption{Detection performance degradation under adversarial perturbations on News, HumanEval, Student Essay, Paper, and Yelp datasets.   
    Each cell shows: perturbed score($\uparrow$) $|$ original score($\uparrow$) $|$ $\downarrow$relative decrease percentage($\downarrow$).
    The \textbf{best} and \underline{second-best} perturbed scores are marked in bold and underlined, respectively.
    }
    \label{table:perturbation}
\end{table*}

\subsubsection{(III) Performance Across Diverse Text Domains.}
As shown in Figure~\ref{figure:datasets}, \ourmethod (both SDC and SDP variants) consistently outperforms all four categories of baselines across five text domains, achieving the broadest coverage as indicated by the outermost curves in each radar chart.
The most significant gains appear in HumanEval and Yelp, where \ourmethod leads by up to 24\% in F1 score. 
This reflects its ability to capture stable sentiment patterns inherent in LLM-generated content, especially in structured code and social reviews.
While margins are smaller for News, Paper, and Essay, our method still demonstrates consistent superiority.
These results confirm that sentiment distribution stability is a robust and transferable signal, enabling \ourmethod to generalize effectively across diverse content types and outperform both business and state-of-the-art detectors.

\begin{figure}[!t]%
    \centering
    \includegraphics[width=\linewidth]{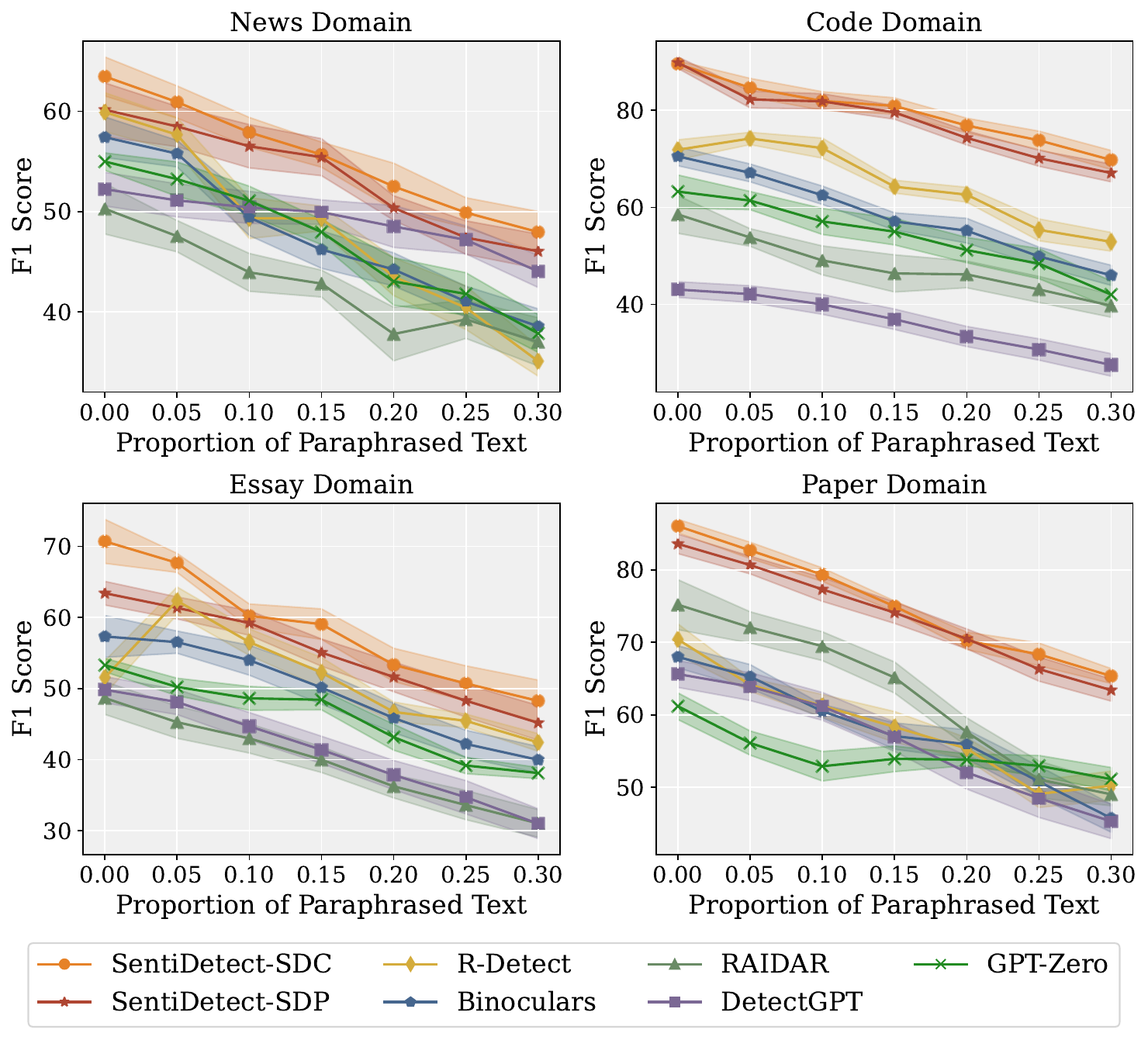}
    \caption{Comparison of paraphrased text on the News, HumanEval, Student Essay, and Paper datasets.}
    \label{figure:paraphrase}
\end{figure}
\subsection{Robustness Analysis and Ablation Study}
\subsubsection{(I) Robustness Against Adversarial Perturbation.}
We evaluate the detection robustness of various methods under adversarial attack~\cite{ren2019generating} using \textit{TextAttack}\footnote{\url{https://github.com/QData/TextAttack}} library to implement it, which simulates practical evasion tactics by applying lexical-level perturbations to deceive classifiers.
Table~\ref{table:perturbation} reports the F1 scores post-attack, alongside original scores and relative degradation percentages. 
Across all five datasets, most baseline detectors, such as GPT-Zero, DetectGPT, and RoBERTa, experience sharp performance drops, often exceeding 70\% loss. 
In contrast, \ourmethod exhibits significantly greater resilience. 
\ourmethod-SDC achieves the highest average detection score under attack, with a comparatively small drop, while \ourmethod-SDP also attains low degradation on average. 
These consistent advantages across datasets underscore that sentiment distribution features remain less sensitive to adversarial text transformations than token-level or syntactic signals.

\subsubsection{(II) Robustness Against Paraphrasing.}
To evaluate the robustness of \ourmethod against paraphrasing attacks, we demonstrate the detection performance of detectors on paraphrased text through \GPTThreeFive in Figure~\ref{figure:paraphrase}.
We control the proportion of changed paraphrasing texts and test the performance at different paraphrasing ratios.
Figure~\ref{figure:paraphrase} shows the experimental results of our method and various baseline methods on the News, HumanEval, Student Essay, and paper datasets.
As the proportion of paraphrased text increases, the performance of all detection methods shows a downward trend, but \ourmethod consistently outperforms all other methods in F1 score and achieves considerable F1 score values on all datasets.
Notably, even if 30\% of the content is replaced, \ourmethod still maintains its strength.
\begin{figure}[!t]%
    \centering
    \includegraphics[width=\linewidth]{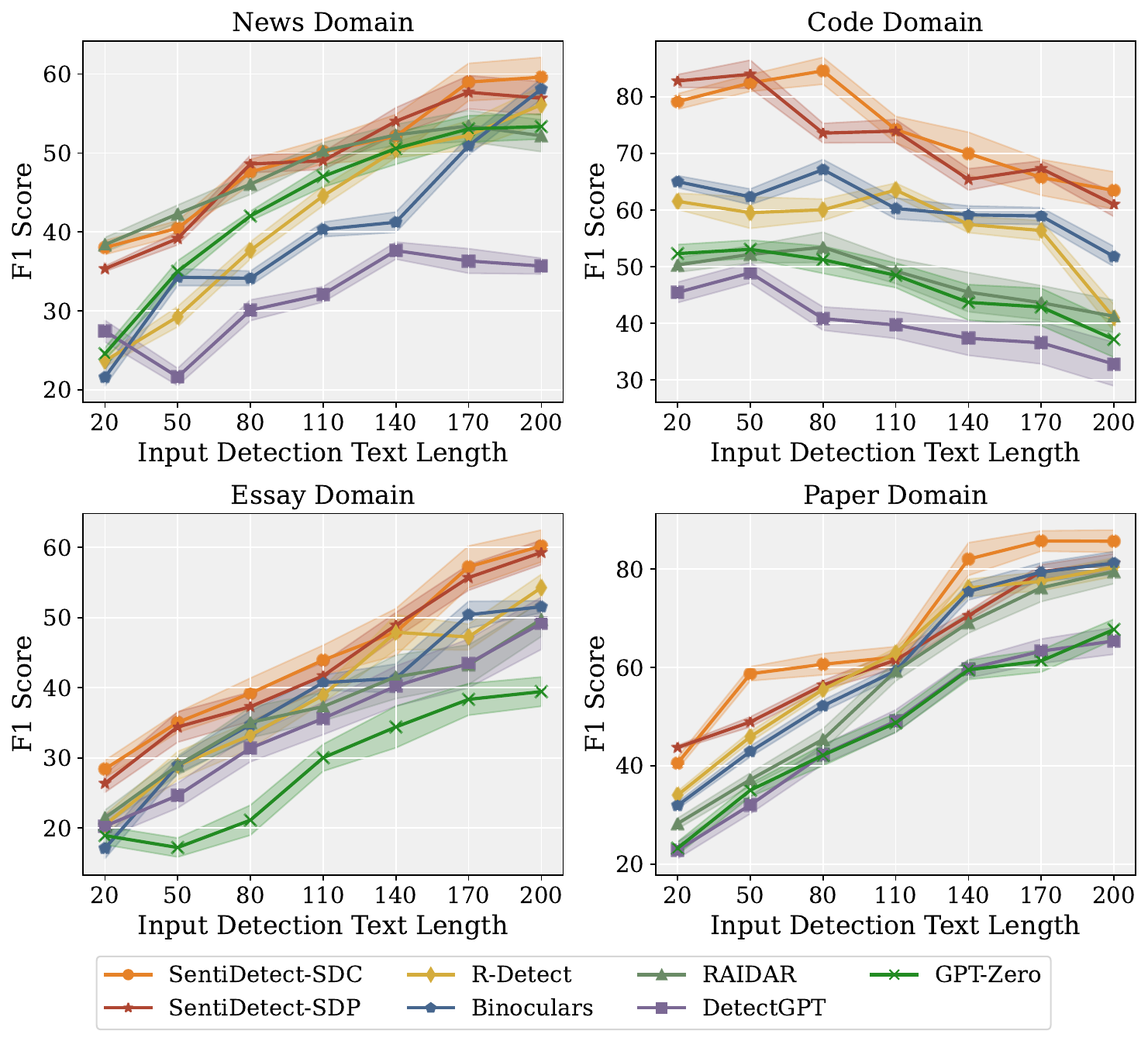}
    \caption{Comparison of varying text length on the News, HumanEval, Student Essay, and Paper datasets.} 
    \label{figure:length}
\end{figure}

\subsubsection{(III) Effect of Input Text Length.}
Due to the sensitivity of detection to the text length, we evaluate the performance of our method on \GPTFour under different input lengths, as shown in Figure~\ref{figure:length}.
For the Yelp and HumanEval datasets, the input texts tend to be short, and thus, it may be necessary to concatenate multiple samples together. 
Each data point represents the average performance for inputs of a specific length, calculated by aggregating samples with a length within ±10 words.
We can see that \ourmethod achieves the highest scores across different lengths and is robust to samples from various fields.
It is also worth noting that many existing detectors show poor performance in situations with short inputs. 
However, even if the input is as short as about 20 words, our method can still achieve superior detection scores, demonstrating its robustness to short text detection.

\textbf{Short Text Detection.} Notably, we observe that zero-shot detectors often require sufficient text length for reliable analysis.
For shorter text samples, advanced language models can more closely approximate human expression patterns, making the distinction more challenging based on limited content.
Aggregating multiple short texts can help capture broader sentiment distribution stability features, thereby improving detection performance.
While longer inputs generally enhance detection performance, the results on HumanEval first increase with text length and then decrease, suggesting that excessively long inputs may not always lead to better results and could even impair detection.

\begin{table}[htbp]
    \centering
    \setlength{\tabcolsep}{4pt}
    \resizebox{\linewidth}{!}{
    \begin{tabular}{c|cccc|cccc}
        \hline
        \hline
        \rowcolor{gray!40}
        & \multicolumn{4}{c}{\textbf{Student Essay}} &  \multicolumn{4}{c}{\textbf{Yelp Review} }\\
        \hline
        \rowcolor{gray!10}
        \textbf{} & $I$=3 & $I$=5 & $I$=7 & $I$=9 & $I$=3 & $I$=5 & $I$=7 & $I$=9 \\
        \hline \hline
        \ourmethod-SDC & 59.03 & 59.56 & 56.62 & 56.60 & 77.02 & 77.58 & 77.74 & 77.63 \\
        \ourmethod-SDP & 56.73 & 57.01 & 56.94 & 56.92 & 80.16 & 80.31 & 80.42 & 80.44 \\
        \hline
    \end{tabular}
    }
    \caption{Effect of the number of LER prompts on the Student Essay and Yelp datasets.}
    \label{tab:rewrite_effect}
\end{table}
\subsubsection{(IV) Number of LER Prompts ($I$).}
We investigate how the number of LER rewriting ($I$) influences detection performance using \GPTFour on the Essay and Yelp datasets.
As shown in Table~\ref{tab:rewrite_effect}, both \ourmethod-SDC and \ourmethod-SDP show performance gains as $I$ increases. 
However, the improvements tend to plateau with larger $I$, indicating diminishing returns. 
These findings suggest that moderate rewriting (e.g., $I{=}5$) is generally effective, while excessive rewriting may offer limited benefits.

\section{Conclusion}
In this paper, we present \ourmethod, a novel model-agnostic framework for detecting LLM-generated text by analyzing sentiment distribution stability, particularly the stability under low-emotional rewriting. 
Based on empirical observations of sentiment divergence, our method achieves superior detection performance across five diverse domains: news articles, programming code, student essays, academic papers, and community comments, leveraging various advanced LLMs.
Experiments demonstrate that \ourmethod outperforms existing state-of-the-art baselines, significantly improving F1 detection scores by over 16\% on \Gemini and over 11\% on \GPTFour.
Furthermore, it also exhibits strong resilience against several adversarial challenges, including paraphrasing attacks, adversarial perturbations, and variable text lengths. 
This work advances the field by bridging the robustness gap in LLM-generated text detection and offers a scalable, cross-domain solution.

\bibliography{aaai2026}

\end{document}